\documentclass{article}

\usepackage{microtype}
\usepackage{graphicx}
\usepackage{subfigure}
\usepackage{booktabs} 
\usepackage{amsmath}
\usepackage{amssymb}
\usepackage{footnote}
\makesavenoteenv{tabular}
\makesavenoteenv{table}
\usepackage{tabularx}

\usepackage[draft]{hyperref}

\usepackage[accepted]{icml2019}

\icmltitlerunning{DeepCABAC: Context-adaptive binary arithmetic coding for deep neural network compression}
\begin{document}

\twocolumn[
\icmltitle{DeepCABAC: Context-adaptive binary arithmetic coding for deep neural network compression}



\icmlsetsymbol{equal}{*}

\begin{icmlauthorlist}
\icmlauthor{Simon Wiedemann}{equal,to}
\icmlauthor{Heiner Kirchhoffer}{equal,to}
\icmlauthor{Stefan Matlage}{equal,to}
\icmlauthor{Paul Haase}{equal,to}
\icmlauthor{Arturo Marban}{to}
\icmlauthor{Talmaj Marinc}{to}
\icmlauthor{David Neumann}{to}
\icmlauthor{Ahmed Osman}{to}
\icmlauthor{Detlev Marpe}{to}
\icmlauthor{Heiko Schwarz}{to}
\icmlauthor{Thomas Wiegand}{to}
\icmlauthor{Wojciech Samek}{to}

\end{icmlauthorlist}

\icmlaffiliation{to}{Department of Video Coding \& Analytics, Fraunhofer Heinrich-Hertz Institut, Berlin, Germany}

\icmlcorrespondingauthor{Wojciech Samek}{wojciech.samek@hhi.fraunhofer.de}

\icmlkeywords{Machine Learning, ICML}

\vskip 0.3in
]



\printAffiliationsAndNotice{\icmlEqualContribution} 

\begin{abstract}
We present DeepCABAC, a novel context-adaptive binary arithmetic coder for compressing deep neural networks. It quantizes each weight parameter by minimizing a weighted rate-distortion function, which implicitly takes the impact of quantization on to the accuracy of the network into account. Subsequently, it compresses the quantized values into a bitstream representation with minimal redundancies.  We show that DeepCABAC is able to reach very high compression ratios across a wide set of different network architectures and datasets.  For instance, we are able to compress by x63.6 the VGG16 ImageNet model with no loss of accuracy, thus being able to represent the entire network with merely 8.7MB.
\end{abstract}

\section{Introduction}
Inspite of their state-of-the-art performance across a wide spectrum of problems \cite{DeepLearning},  deep neural networks have the well-known caveat that most often they have high memory complexity. This does not only imply high storage capacities as a requirement, but also high energy resources and slower runtimes for execution \cite{Horowitz, Efficient_DNN_processing, Custom_hardware_DNN_survey}. This greatly limits their adoption in industrial applications or their deployment into resource constrained devices. Moreover, this also difficults their transmission into communication channels with limited capacity, which becomes an obstacle for distributed training scenarios such as in federated learning \cite{federated_learning, SatArXiv18, SatArXiv19}.  

As a reaction, a plethora of work has been published on the topic of deep neural network compression \cite{DLC_survey, DLC_survey2}. From all different proposed methods, sparsification followed by weight quantization  and entropy coding arguably belong to the set of most popular approaches, since very high compression ratios can be achieved under such paradigm  \cite{deep_compression, BayesianCompression, Entropy_constrained_DNN, Simon_lossless_dnn_compression1}.  Whereas much of research has focused on the sparsification part, a substantially less amount have focused on improving the later two steps. In fact, most of the proposed (post-sparsity) compression algorithms come with at least one of the following caveats: 1) they decouple the quantization procedure from the subsequent lossless compression algorithm, 2) ignore correlations between the parameters and 3) apply a lossless compression algorithm that produce a bitstream with more redundancies than principally needed (e.g. scalar Huffman coding). Moreover, some of the proposed compression algorithms do also not take the impact of quantization on to the accuracy of the network into account.

In this work we present DeepCABAC, a compression algorithm that overcomes all of the above limitations. It is based on applying a context-adaptive binary arithmetic coder (CABAC) on to the quantized parameters, which is the state-of-the-art for lossless compression. It also couples the quantization procedure with CABAC by minimizing a rate-distortion cost function where the rate explicitly measures the bit-size of the network parameters as determined by CABAC. Moreover, it implicitly takes the impact of quantization on to the networks accuracy into account by weighting the distortion with a term that measures the ``robustness'' of the networks parameter. In our experiments we show that we can significantly boost the compression performance of a wide set of pre-sparsified network architectures, consequently achieving new state-of-the-art results for the VGG16 model.

\section{CABAC}
Context-adaptive binary arithmetic coding (CABAC) is a form of lossless coding which was originally designed for the video compression standard H.264/AVC \cite{CABAC}, but it is also an integral part of its successor H.265/HEVC.  CABAC does not only offer high flexibility of adaptation, but also a highly efficient implementation, thus attaining higher compression performance as well as throughputs compared to other entropy coding methods \cite{CABAC_efficient}. In short, it applies three powerful coding techniques: 1) Firstly, it binarizes the data to be encoded. That is, it predefines a series of binary decisions (also called bins) under which each data element (or symbol) will be uniquely identified. 2) Then, it assigns a binary probability model to each bin (also named context model) which is updated on-the-fly by the local statistics of the data. This enables CABAC with a high degree of adaptation to different data distributions. 3) Finally, it employs an arithmetic coder in order to optimally and efficiently code each bin, based on the respective context model. To recall, arithmetic coding is a form of entropy coding which encodes entire strings of symbols into a single integer value. It is well-known to outperform other coding techniques such as the Huffman code \cite{Huffman} with regards to both, compactness of the data representation and coding efficiency  \cite{arithmetic_coding}. 

Due to the above reasons, we chose CABAC as our lossless compression method and adapted it for the task of neural network compression. 

\subsection{Binarization on deep neural networks}
Inspired by a prior analysis on the empirical weight distribution of different neural network architectures, we adopted the following bianrization procedure. Given a quantized weight tensor in its matrix form\footnote{For fully-connected layers this is trivial. For convolutional layers we converted them into their respective matrix form according to \cite{NVIDIA_convolution}.}, DeepCABAC scans the weight elements in row-major order\footnote{From left to right, up to down.} and encodes each quantized weight element value by: 1) firstly determining if the weight element is a significant element or not. That is, each weight element is assigned with a bit which determines if the element is 0 or not. This bit is then encoded using a binary arithmetic coder, according to its respective context model. The context model is initially set to 0.5 (thus, 50\% probability that a weight element is 0 or not), but will automatically be adapted to the local statistics of the weight parameters as DeepCABAC encodes more elements.
2) Then, if the element is not 0, the sign bit is analogously encoded, according to its respective context model.
3) Subsequently, a series of bits are analogously encoded, which determine if the element is greater than $1,2,...,n \in \mathbb{N}$. The number $n$ becomes a hyperparameter for the encoder.
4) Finally, the reminder is encoded using a fixed-length binary code. 

\begin{figure}[t]
\centering
\includegraphics[width=\columnwidth,clip,keepaspectratio]{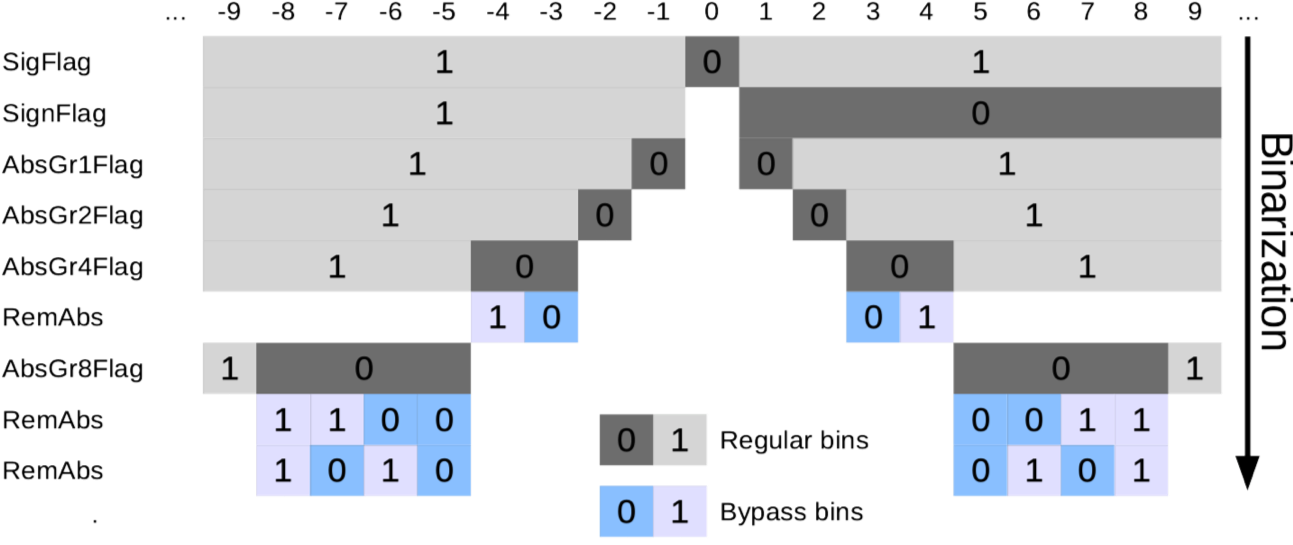}
\caption{DeepCABAC binarization of neural networks. It encodes each weight element by performing the following steps: 1) encodes a bit named \textit{sigflag} which determines if the weight is a significant element or not (in other words, if its 0 or not). 2) If its not 0, then the sign bit, \textit{signflag}, is encoded. 3) Subsequently, a series of bits are encoded, which indicate if the weight value is greater equal than $1, 2, ..., n \in \mathbb{N}$ (the so called  \textit{AbsGr(n)Flag}). 4) Finally, the reminder is encoded. The grey bits (also named regular bins) represent bits that are encoded using an arithmetic coder according to a context model. The other bits, the so called bypass bins, are encoded in fixed-point form. The decoder is analogous.}
\label{Fig: CABAC}
\end{figure}

The decoding process is performed analogously. An example scheme of the binarization procedure is depicted in figure \ref{Fig: CABAC}.

\section{Weighted rate-distortion function}
Before we apply CABAC, we have to firstly quantize the weight parameters of the network. We do this by minimizing a generalised form of a rate-distortion function. Namely, we quantize each weight parameter $w_i$ to the quantization point $q_{k^*}$ that minimizes the cost function
\begin{equation}
w_i \rightarrow q_{k^*} = \min_k \eta_i (w_i - q_{k})^2 + \lambda R_{ik} 
\label{Eq: WRD function}
\end{equation}
where $R_{ik}$ is the bit-size of the quantization point $q_{k}$ as determined by DeepCABAC, and $\lambda$ is the lagrangian multiplier that specifies the desired trade-offs between the bit-size and distortion incurred by the quantization. Notice, how the bit-size $R_{ik}$ now also depends on the index $i$ of the weight to be encoded. This is due to the context-adaptive models which update their probabilities as new elements are being encoded, thus being different for each weight $w_i$ and consequently the bit-size of each quantization point $q_k$.

Moreover, \eqref{Eq: WRD function} introduces a weight-specific parameter $\eta_i$ which takes into account the relative impact that the distortion of a particular weight inccurs on to the accuracy of the network. In this work we take a bayesian approach in order to estimate this parameter. Namely, we assume a gaussian prior for each weight parameter and apply scalable bayesian techniques in order to estimate their sufficient statistics \cite{variational_dropout, VD_sparsifies, BayesianCompression}. As a result, we attain a mean and standard deviation value for each weight parameter of the network, where the former can be interpreted as its (new) value and the later as a measure of its ``robustness''. Thus, when quantizing each $w_i$, we set $\eta_i = 1/\sigma_i^2$ in \eqref{Eq: WRD function}, with  $\sigma_i$ being the respective standard deviation. This is also theoretically motivated, since \cite{CLP_achile} established a connection between the variances and the diagonal elements of the fisher information matrix.

In order to minimize \eqref{Eq: WRD function}, we also need to define a set of quantization points $q_{k}$. We chose them to be equidistant to each other with a particular distance $\Delta \in \mathbb{R}$, namely, 
\begin{equation}
q_{k} = \Delta I_k, \quad \Delta =\frac{2|w_{\max}|}{ \frac{2|w_{\max}|}{\sigma_{\min}} + S},  \quad S,I_k \in \mathbb{Z}
\label{Eq: quantization points}
\end{equation}
where $\sigma_{\min}$ is the smallest standard deviation and $w_{\max}$ the parameter with highest magnitude value. $S$ is then a hyperparameter, which controls the ``coarseness'' of the quantization points. By selecting $\Delta$ in such a manner we ensure that the quantisation points lie within the range of the standard deviation of each weight parameter, in particular for values $S \geq 0$. Moreover, by constraining them to be equidistant to each other we encourage fixed-point representations, which can be exploited in order to perform inference with lower complexity \cite{QNNPACK, TFlite}.

\section{Experiments}

\begin{table}[!t]
\centering
\caption{Compression ratios achieved when combining DeepCABAC with a sparsification method. In parenthesis are the results from previous work, where $^1$\cite{deep_compression} and $^2$\cite{BayesianCompression}.}
\vspace{0.1in}

\resizebox{\columnwidth}{!}{%
\begin{tabular}{|p{1.3cm}|p{1.4cm}|p{1.0cm}|p{0.9cm}|p{0.95cm}|p{1.0cm}|p{1.1cm}|}
\hline
 Models  & Dataset & Org.acc. (top1)  [\%] & Org. size & Spars. $\frac{|w \neq 0|}{|w|}$ \vspace{0.05in} [\%] & Comp. ratio [\%] & Acc. (top1) [\%]\\
\hline
\hline
VGG16 & ImageNet & 69.43 & 553.43 MB  & 9.85 (7.5$^1$)& \textbf{1.57} (2.05$^1$) & \textbf{69.43} (68.83$^1$)  \\
\hline
ResNet50 & ImageNet & 76.13 & 102.23 MB & 25.40 (29.0$^1$)& \textbf{5.95} (5.95$^1$) & \textbf{74.12} (76.15$^1$)  \\
\hline
Mobile-Net-v1 & ImageNet & 70.69 & 16.93 MB  & 50.73 & \textbf{12.7}  & \textbf{66.18} \\
\hline
Small-VGG16 & CIFAR10 & 91.35 & 59.9 MB & 7.57 (5.5$^2$) & \textbf{1.6} (0.86$^2$) & \textbf{91.00} (90.8$^2$)\\
\hline
LeNet5 & MNIST & 99.22 & 1722 KB & 1.90 (8.0$^1$) (0.6$^2$) & \textbf{0.72} (2.55$^1$) (0.13$^2$) & \textbf{99.16}  (99.26$^1$) (99.00$^2$) \\
\hline
LeNet-300-100 & MNIST & 98.29 & 1066 KB & 9.05 (8.0$^1$) (2.2$^2$) & \textbf{1.82} (2.49$^1$) (0.88$^2$) & \textbf{98.08}  (98.42$^1$) (98.20$^2$) \\
\hline
\hline
FCAE & CIFAR10 & 30.14 PSNR & 304.72 KB & 55.69 & \textbf{16.15} & \textbf{30.09} PSNR \\
\hline
\end{tabular}
}
\label{Tbl: experimental results}
\end{table}
\footnotetext{\url{https://github.com/slychief/ismir2018_tutorial/tree/master/metadata}}

We applied DeepCABAC on the set of models described in the evaluation framework \cite{MPEG_eval_framework} of the MPEG call on neural network representations \cite{MPEG_nncompression_call}. This includes the VGG16, ResNet50 and MobileNet-v1 models and a fully-convolutional autoencoder pretrained on a task of end-to-end image compression (which we named \textit{FCAE}). In addition, we also applied DeepCABAC on the LeNet-300-100 and LeNet5 models and on a smaller version of VGG16\footnote{\url{http://torch.ch/blog/2015/07/30/cifar.html}} model (which we named \textit{Small-VGG16}). 

We aplied the variational sparsification method introduced in \cite{VD_sparsifies} on to the LeNet-300-100, LeNet5, Small-VGG16, FCAE and MobileNet-v1 models. However, due to the high training complexity that this method requires, we adopted a slightly different approach for the VGG16 and ResNet50. Namely, we firstly sparsified them by applying the iterative algorithm \cite{Learning_Weight_and_Connections}, and subsequently applied method \cite{VD_sparsifies} but only for estimating the variances of the distributions (thus, fixing the mean values during training). After sparsification, we applied DeepCABAC on to the weight parameters of each layer separately, excluding biases and normalization parameters. Since the compression result can be sensitive to the parameter $S$ in \eqref{Eq: quantization points}, we probed the compression performance for all $S \in \{0,1,...,256\}$ and selected the best performing model.

The resulting sparsities as well as the compression ratios are displayed in table \ref{Tbl: experimental results}. Notice that for most networks we are not able to reproduce the sparsity ratios reported in the literature. In addition, we did not perform any fine-tuning after compression, thus having a particularly challenging setup for achieving good post-sparsity compression ratios. Nevertheless, in-spite of these two disadvantages, DeepCABAC is able to significantly compress further the models, boosting the compression by 74\% ($\pm$ 8\%) on average and consequently achieving compression results comparable to the current state-of-the-art. Moreover, we are able to compress by x63.6 (1.57\%) the VGG16 model without loss of accuracy, thus reaching a new state-of-the-art benchmark. 

\section{Conclusion}
We show that one can boost significantly the compression gains if one applies state-of-the-art coding techniques on to pre-sparsified deep neural networks. In particular, our proposed coding scheme, DeepCABAC, is able to increase the compression rates of pre-sparsified networks by 74\% on average, attaining as such compression ratios comparable (or sometimes higher) to the current state-of-the-art. In future work we will benchmark DeepCABAC also on non-sparsified networks, as well as apply it in the context of distributed learning scenarios where memory complexity is critical (e.g. in federated learning).

\clearpage
\bibliographystyle{icml2019}
\bibliography{../../References}

\end{document}